\documentclass{article}

\usepackage{arxiv}

\usepackage{cite}
\usepackage{amsmath,amssymb,amsfonts}
\usepackage{bm}
\usepackage{mathtools}
\usepackage{graphicx}
\usepackage{textcomp}
\usepackage{xcolor}
\usepackage{booktabs}
\usepackage{multirow}
\usepackage{array}
\usepackage{siunitx}
\usepackage{enumitem}
 \usepackage{url}
\usepackage{tikz}
\usepackage{orcidlink}

\usetikzlibrary{arrows.meta,positioning,shapes.geometric,fit,calc}

\usepackage[ruled,vlined]{algorithm2e}

\hypersetup{
    colorlinks=true,
    linkcolor=blue,
    citecolor=blue,
    urlcolor=blue
}

\graphicspath{{figures/}}

\newcommand{\R}{\mathbb{R}}

\newcommand{\diag}{\mathrm{diag}}

\newcommand{\argmax}{\operatornamewithlimits{arg\,max}}

\newcommand{\AuthorBio}[3]{%
\begin{figure}[h]
    \centering
    \begin{minipage}{0.18\textwidth}
        \includegraphics[
            width=3cm,
            height=4cm,
            keepaspectratio,
            clip
        ]{#1}
    \end{minipage}
    \hfill
    \begin{minipage}{0.78\textwidth}
        \textbf{#2} #3
    \end{minipage}
\end{figure}
}

\def\BibTeX{{\rm B\kern-.05em{\sc i\kern-.025em b}\kern-.08em
    T\kern-.1667em\lower.7ex\hbox{E}\kern-.125emX}}

\title{BayesFusion--SDF: Probabilistic Signed Distance Fusion with View Planning on CPU}

\author{
Soumya Mazumdar~\orcidlink{0009-0006-3521-9557} \\
Department of Computer Science and Business Systems \\
Gargi Memorial Institute of Technology \\
Baruipur, Kolkata, West Bengal 700144, India \\
\texttt{reachme@soumyamazumdar.com} \\
\And
Vineet Kumar Rakesh~\orcidlink{0009-0000-7102-6564} \\
Computer and Informatics Group, Variable Energy Cyclotron Centre \\
1/AF, Bidhannagar, Kolkata, West Bengal 700064, India \\
\texttt{vineet@vecc.gov.in} \\
\And
Tapas Samanta~\orcidlink{0000-0003-0521-0747} \\
Computer and Informatics Group, Variable Energy Cyclotron Centre \\
1/AF, Bidhannagar, Kolkata, West Bengal 700064, India \\
Engineering Sciences, Homi Bhabha National Institute \\
Training School Complex, Anushaktinagar, Mumbai, Maharashtra 400094, India \\
\texttt{tsamanta@vecc.gov.in}
}

\begin{document}
\maketitle

\begin{abstract}
Key part of robotics, augmented reality, and digital inspection is dense 3D reconstruction from depth observations. Traditional volumetric fusion techniques, including truncated signed distance functions (TSDF), enable efficient and deterministic geometry reconstruction; however, they depend on heuristic weighting and fail to transparently convey uncertainty in a systematic way. Recent neural implicit methods, on the other hand, get very high fidelity but usually need a lot of GPU power for optimization and aren't very easy to understand for making decisions later on. This work presents BayesFusion SDF, a CPU-centric probabilistic signed distance fusion framework that conceptualizes geometry as a sparse Gaussian random field with a defined posterior distribution over voxel distances. First, a rough TSDF reconstruction is used to create an adaptive narrow-band domain. Then, depth observations are combined using a heteroscedastic Bayesian formulation that is solved using sparse linear algebra and preconditioned conjugate gradients. Randomized diagonal estimators are a quick way to get an idea of posterior uncertainty. This makes it possible to extract surfaces and plan the next best view while taking into account uncertainty. Tests on a controlled ablation scene and a CO3D object sequence show that the new method is more accurate geometrically than TSDF baselines and gives useful estimates of uncertainty for active sensing. The proposed formulation provides a clear and easy-to-use alternative to GPU-heavy neural reconstruction methods while still being able to be understood in a probabilistic way and acting in a predictable way. GitHub: \href{https://mazumdarsoumya.github.io/BayesFusionSDF}{https://mazumdarsoumya.github.io/BayesFusionSDF}
\end{abstract}


\section{Introduction}
Dense 3D reconstruction from multi-view imagery and depth sensing is a fundamental challenge in computer vision, facilitating applications in augmented and virtual reality, robotics, autonomous navigation, cultural heritage preservation, and digital content creation. The Classical Volumetric Fusion Techniques (CVFT) and the Truncated Signed Distance Field (TSDF) can be observed as mathematical representation which utilized the Iso-surface algorithms with Marching Cubes~\cite{curless1996volumetric,lorensen1987marching}, to extract the 3D surface value (isovalue). The real-time systems such as KinectFusion and the scalable version of that pipelines can be used as affordable sensors to make such strong and high quality reconstructions~\cite{newcombe2011kinectfusion,niessner2013voxelhashing,kahler2015infinitam,oleynikova2017voxblox,zhou2018open3d}. Although those methods are extremely useful and effective, those methods use heuristic weighting schemes and deterministic fusion rules which means that those methods do not give principled uncertainty for the help in the tasks like confidence aware perception, view planning or decision making for safety. 

The latest advancements in the representation of neural implicit have enhanced the reconstructions fidelity and the rendering realism. Neural Radiance Fields (NeRF) and Neural Signed Distance Functions (SDFs) learn continuous scene changing from images achieving the results of state-of-the-art (SOTA) in both geometry and appearance modeling~\cite{yariv2021volsdf,wang2021neus}. Also, hybrid methods such as Gaussian Splatting make rendering faster and better~\cite{kerbl20233dgs,gsurf2024,gsfusion2024} but these kinds of methods are relied on optimization which requires sufficient time on GPUs, requiring additional training time and might not show uncertainty in a way that fits well with traditional mapping or robotics pipelines. Hence, there is a gap between High-Fidelity Neural Surface Reconstruction and the geometry pipeline for the real world which requires less resources. Previous works in the probabilistic fusion and the uncertainty aware SLAM indicates the integration of sensor noise models and the spatial correlations in enhancing robustness and information driven exploration~\cite{rosinol2022probvolfusion,uncertaintyvislam2024,vdbgpdf2024}. Sparse precision formulation utilizes Gaussian Markov random fields (GMRFs) for appealing the exploiting inference through iterative solvers~\cite{rue2005gmrfs,saad2003iterative}, rendering appropriate for CPU only deployment.

Uncertainty-aware geometry also makes active perception possible. This includes next-best-view (NBV) planning, in which sensor poses are chosen to get the most information possible. Information-theoretic and learning-based next-best-view (NBV) methods have been shown to work well at reducing reconstruction ambiguity and improving coverage~\cite{banta2000nbv,nbvprob2014,almadhoun2019guided,peralta2020scanrl}. Nonetheless, incorporating NBV objectives directly into volumetric fusion pipelines is difficult without a clear probabilistic representation of geometry uncertainty.

This paper presents \emph{BayesFusion SDF}, a probabilistic signed distance fusion framework intended for CPU-first deployment, which enhances traditional TSDF reconstruction through Bayesian inference and uncertainty estimation. The method uses a TSDF-based geometric initialization and a sparse Gaussian random field model. This makes it possible to scale inference using sparse linear algebra and preconditioned iterative solvers. Using randomized diagonal approximation techniques, we can estimate posterior uncertainty, which makes it easy to do calculations close to the reconstructed surface. The uncertainty representation that comes out of this is used directly for active view planning with the goal of reducing expected variance. The framework keeps the determinism, controllability, and simplicity of classical volumetric fusion while making uncertainty a first-class output that can be used in vision and robotics.

The main points of the contributions are as follows:
\begin{itemize}
    \item A probabilistic distance fusion framework for CPUs for sparse Gaussian random field formulation.
    \item A method for estimating uncertainty with large voxel domains using randomized probe diagonal approximation. 
    \item An uncertainty-driven next-best-view planning formulation integrating with volumetric reconstruction. 
    \item Experimental results show better geometric quality and useful estimates than TSDF baselines.
\end{itemize}

\section{Related Work}

The Truncated Signed Distance Fields (TSDFs) is the most used method for the densely 3D reconstruction for predictable and fast computing. The groundbreaking work of Curless and Levoy introduced the signed distance integration for range in image fusion~\cite{curless1996volumetric} that made surface extraction efficient and it is later on followed by Marching Cubes \cite{lorensen1987marching}. KinectFusion~\cite{newcombe2011kinectfusion} which is a real-time dense 3D reconstruction system that works online with cheap RGB-D sensors. Multiple research enhancement has been occurred for scalability and memory efficiency using voxel hashing, hierarchical data structures, and sparse representations \cite{niessner2013voxelhashing,kahler2015infinitam,oleynikova2017voxblox,zhou2018open3d} but still this account as less useful in terms of perception and planning.

NeRF and SDF receive scenes continuously directly from photos and permitting high fidelity photorealistic rendering reconstruction~\cite{yariv2021volsdf,wang2021neus}. Recent studies based on Gaussian Splating are efficient to render real-time while preserving high quality reconstruction~\cite{kerbl20233dgs,gsurf2024,gsfusion2024}, but certain limitations can be observed while implementing characterized scenarios due to resource constraints or robotics focus as the neural reconstruction methods requires huge memory and huge intensive GPU optimization. Alternative probabilistic neural methods can be implemented but usually rely on sampling oriented inferencing. This leads to the need of methods that can maintain interpretability and computational predictability incorporating principled uncertainty modeling.

Multiple probabilistic volumetric reconstruction methods capture noise and spatial correlations. Thus, various studies have been conducted to examine Bayesian fusion, uncertainty-aware SLAM, and probabilistic occupancy or distance representations to enhance robustness and uncertainty awareness~\cite{rosinol2022probvolfusion,uncertaintyvislam2024}. Further studies on geometry models and Gaussian processes provide theoretical concepts of sound uncertainty estimations however appropriate computational challenges should be covered in extensive context~\cite{vdbgpdf2024}. The derived formula of Gaussian Markov random fields (GMRFs) utilizing sparse precision matrices and local structural facilitating efficient inference using iterative solvers can be used as a solution~\cite{rue2005gmrfs,saad2003iterative}. This has the predictable convergence properties, understandable uncertainty, and CPU based implementations.

The next-best-view (NBV) methods are geometric concentrated and measures heuristic~\cite{banta2000nbv}, while still requires sufficient development in probabilistic reasoning, information theoretic aims, and learning-based policies~\cite{nbvprob2014,almadhoun2019guided,peralta2020scanrl}. In addition, the relation between the active learning and experimental design has been studied specially in perspective selection to reduce posterior uncertainty in spatial models~\cite{mackay1992active,cohn1996active}. Learnt representation view planning merged with neural reconstruction techniques to inherit the computational complexity and hardware constraints neural training pipelines utilizing for computational cost.

The comparison between the traditional TSDF fusion, neural implicit reconstruction and proposed probabilistic formulation can be observed in \ref{tab:comparison}. Classical methods can provide efficiency and predictability behaviours but not the principled uncertainty while neural approaches yield higher fidelity but require intense GPU. Thus, the proposed technique blend deterministic volumetric fusion with probabilistic modeling while deploying in CPU only devices

\begin{table*}[h]
\centering
\caption{Comparison of representative dense reconstruction paradigms.}
\label{tab:comparison}
\setlength{\tabcolsep}{4pt}
\begin{tabular}{p{2.7cm}p{3.4cm}p{3.8cm}p{5.2cm}}
\toprule
\textbf{Aspect} & \textbf{Traditional TSDF Fusion} & \textbf{Neural Implicit Reconstruction} & \textbf{Proposed Probabilistic SDF (BayesFusion)} \\
\midrule

Representation &
Discrete voxel SDF &
Continuous neural field &
Probabilistic voxel SDF (GRF) \\

Computation &
CPU/GPU, real-time feasible &
GPU, training-intensive &
CPU-first sparse linear algebra \\

Uncertainty Estimation &
Heuristic weights &
Limited / indirect &
Explicit posterior variance \\

Determinism &
High &
Optimization-dependent &
High (solver-based) \\

Training Requirement &
None &
Required &
None \\

Scalability &
Good with sparse structures &
Heavy for large scenes &
Sparse hierarchical domain \\

Robotics Integration &
Mature &
Limited &
Direct compatibility \\

Active View Planning &
Not native &
Emerging &
Native (variance reduction) \\

Hardware Dependence &
Flexible &
GPU-centric &
CPU-oriented \\

Interpretability &
High &
Limited &
High probabilistic interpretability \\

\bottomrule
\end{tabular}
\end{table*}

The proposed approach implements the complementary position of the deterministic volumetric fusion and the neural implicit reconstruction. Comparing with the classical TSDF methods and the introduced method principled probabilistic inference and uncertainty estimation preserving computational efficiency and deployment simplicity. Additionally, the proposed framework does not require GPU based training and provides explicit posterior uncertainty while integrating naturally planning robotics pipelines by combining TSDF initialization with Gaussian random field (GRF).

\section{Method}
\subsection{Problem Setup and Notation}
A sequence of frames is indexed by $t=1,\dots,T$.
Each frame provides (optionally):
\begin{itemize}[leftmargin=1.2em]
  \item a depth map $D_t(\bm{u})$ (or depth prediction) at pixel $\bm{u}$,
  \item a pose $T_{w\leftarrow c_t}\in SE(3)$ mapping camera $c_t$ to world $w$,
  \item intrinsics $\mathbf{K}$.
\end{itemize}

Let $\phi:\R^3\to\R$ denote the signed distance field (SDF), with surface $\mathcal{S}=\{\bm{x}:\phi(\bm{x})=0\}$.
The field $\phi$ is discretized on an adaptive sparse voxel hierarchy resulting in unknown vector $\bm{x}\in\R^{N}$ of SDF values at active voxel nodes.
Noisy distance observations are denoted by $y_i$ with corresponding model row $\bm{a}_i^\top$ such that $y_i \approx \bm{a}_i^\top \bm{x}$.
Each observation has (possibly heteroscedastic) noise variance $\sigma_i^2$.

\subsection{Pipeline Overview}
\begin{figure}[h]
\centering
\begin{tikzpicture}[
  node distance=6mm and 6mm,
  box/.style={draw, rounded corners, align=center, inner sep=3pt, font=\small},
  arrow/.style={-Latex, thick}
]
\node[box] (A) {Depth maps\\ + Poses};
\node[box, right=of A] (B) {Preprocess\\ (filter,\\ noise model)};
\node[box, right=of B] (C) {Coarse TSDF\\ bootstrap};
\node[box, below=of C] (D) {Sparse voxel\\ hierarchy};
\node[box, left=of D] (E) {Probabilistic SDF\\ (sparse GRF)};
\node[box, below=of E] (F) {PCG solve\\ (MAP SDF)};
\node[box, right=of F] (G) {Approx. variance\\ (probes)};
\node[box, below=of F] (H) {Surface extraction\\ (MC/DC)};
\node[box, below=of G] (I) {NBV planning\\ (variance reduction)};

\draw[arrow] (A) -- (B);
\draw[arrow] (B) -- (C);
\draw[arrow] (C) -- (D);
\draw[arrow] (D) -- (E);
\draw[arrow] (E) -- (F);
\draw[arrow] (E) -- (G);
\draw[arrow] (F) -- (H);
\draw[arrow] (G) -- (I);
\end{tikzpicture}
\vspace{-0.6em}
\caption{Semantic diagram of BayesFusion--SDF}
\label{fig:pipeline}
\end{figure}
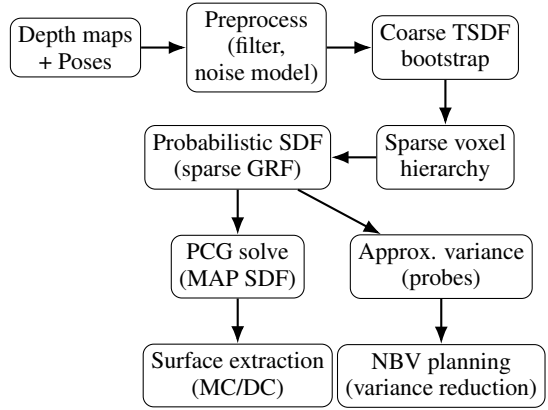

\ref{fig:pipeline} gives a semantic view of the full CPU-only pipeline.

\subsection{TSDF Bootstrap and Active Region Selection}
A coarse TSDF volume is integrated using a standard weighted average fusion rule~\cite{curless1996volumetric,zhou2018open3d}.
This provides a rough surface estimate $\hat{\mathcal{S}}_0$ and a narrow-band region $\mathcal{B}$ near the surface where probabilistic refinement is most valuable.
Let $\tau$ be the truncation distance. Voxels whose TSDF magnitude is below the threshold are activated:
\begin{equation}
\mathcal{B} \;=\;\{ \bm{v} : |\mathrm{TSDF}(\bm{v})| \le \alpha\tau \},\quad \alpha\in[1,3].
\end{equation}

\subsection{Adaptive Sparse Voxel Hierarchy}
The field is represented with a sparse voxel hierarchy (octree / VDB / hashed blocks)~\cite{niessner2013voxelhashing,oleynikova2017voxblox,vdbfusion2022}.
Each leaf stores SDF unknown(s) at voxel centers (or corners), and neighbor connectivity is defined by a local stencil (6/18/26 neighbors).
Let $\bm{x}\in\R^N$ be the vector of unknown SDF values for all active nodes.

\subsection{Observation Model (Depth to SDF Samples)}
Each depth pixel $\bm{u}$ yields a 3D point:
\begin{equation}
\bm{p}_{t}(\bm{u}) \;=\; T_{w\leftarrow c_t} \Big( D_t(\bm{u}) \,\mathbf{K}^{-1}\tilde{\bm{u}} \Big),
\end{equation}
where $\tilde{\bm{u}}$ is the homogeneous pixel coordinate.
Samples are generated along the camera ray to obtain signed distance observations to the surface. A common approximation is projective SDF:
\begin{equation}
y_i \;=\; \mathrm{clamp}\big( \langle \bm{n}, (\bm{x}_i-\bm{p}_i)\rangle,\, -\tau,\, \tau \big),
\end{equation}
where $\bm{x}_i$ is a voxel location and $\bm{n}$ approximates a local normal.
For each observation, a (possibly interpolated) linear relation is used:
\begin{equation}
y_i \approx \bm{a}_i^\top \bm{x}, \quad \bm{a}_i \in \R^N,
\end{equation}
where $\bm{a}_i$ has up to 8 non-zeros for trilinear interpolation.
The noise variance $\sigma_i^2$ is assigned using a depth-dependent sensor model and a pose uncertainty approximation:
\begin{equation}
\sigma_i^2 \;=\; \sigma^2_{\text{depth}}(D_t(\bm{u})) \;+\; \sigma^2_{\text{pose}}(t) \;+\; \sigma^2_{\text{model}}.
\end{equation}

\subsection{Sparse GRF Prior (GMRF) and Bayesian Fusion}
The unknown field $\bm{x}$ is modeled as a Gaussian random field with a sparse precision prior (GMRF)~\cite{rue2005gmrfs}:
\begin{equation}
p(\bm{x}) \propto \exp\Big(-\tfrac{1}{2} \bm{x}^\top \mathbf{Q}_0 \bm{x}\Big),
\end{equation}
where $\mathbf{Q}_0$ is sparse and encodes smoothness, e.g., a weighted Laplacian with an (optional) boundary anchoring term:
\begin{equation}
\bm{x}^\top \mathbf{Q}_0 \bm{x}
=
\lambda \sum_{(i,j)\in \mathcal{E}} w_{ij} (x_i - x_j)^2
+
\lambda_b \sum_{i\in \partial\mathcal{B}} (x_i - x_i^{\text{TSDF}})^2.
\end{equation}
Observations satisfy:
\begin{equation}
p(\bm{y}\mid \bm{x}) \;=\; \prod_{i=1}^M \mathcal{N}(y_i \mid \bm{a}_i^\top \bm{x},\, \sigma_i^2).
\end{equation}
Let $\mathbf{A}\in\R^{M\times N}$ stack $\bm{a}_i^\top$, $\bm{y}\in\R^M$ stack $y_i$, and $\mathbf{W}=\diag(\sigma_1^{-2},\dots,\sigma_M^{-2})$.
The posterior is Gaussian:
\begin{equation}
p(\bm{x}\mid \bm{y}) = \mathcal{N}(\bm{\mu}, \mathbf{\Sigma}),
\end{equation}
with precision and mean:
\begin{align}
\mathbf{Q} &= \mathbf{Q}_0 + \mathbf{A}^\top \mathbf{W}\mathbf{A}, \label{eq:precision}\\
\bm{\mu} &= \mathbf{Q}^{-1}(\mathbf{A}^\top \mathbf{W}\bm{y} + \mathbf{b}_0). \label{eq:mean}
\end{align}
The MAP estimate solves $\mathbf{Q}\bm{\mu}=\bm{h}$ with $\bm{h}=\mathbf{A}^\top \mathbf{W}\bm{y} + \mathbf{b}_0$, using PCG~\cite{saad2003iterative}.

\subsection{Approximate Posterior Variance}
The diagonal variance $\diag(\mathbf{Q}^{-1})$ is estimated using randomized probe vectors~\cite{hutchinson1990trace,bekas2007diag}.
Let $\bm{z}^{(k)}\in\{-1,+1\}^{N}$ be Rademacher probes and solve $\mathbf{Q}\bm{u}^{(k)} = \bm{z}^{(k)}$.
An unbiased diagonal estimator is:
\begin{equation}
\widehat{\diag(\mathbf{Q}^{-1})}
=
\frac{1}{K}\sum_{k=1}^{K} \bm{z}^{(k)} \odot \bm{u}^{(k)}.
\end{equation}

\subsection{Surface Extraction and NBV Planning}
Given $\bm{\mu}$ at voxel nodes, mesh $\hat{\mathcal{S}}$ is extracted using Marching Cubes~\cite{lorensen1987marching} or Dual Contouring~\cite{ju2002dual}. A next view $v^\star$ can be selected from candidates $\mathcal{V}$ by maximizing an expected variance-reduction utility on a narrow band around $\phi(\bm{x})=0$:
\begin{equation}
v^\star = \argmax_{v\in\mathcal{V}} \; U(v),\quad
U(v) \approx \sum_{i\in\Omega_\epsilon} \big(\hat{s}_i - \hat{s}'_i\big).
\end{equation}

\begin{algorithm}[h]
\DontPrintSemicolon
\KwIn{Depth frames ${D_t}*{t=1}^{T}$, camera poses ${T*{w\leftarrow c_t}}_{t=1}^{T}$, intrinsics $\mathbf{K}$}
\KwOut{Mesh $\hat{\mathcal{S}}$, uncertainty $\hat{\mathbf{s}}$, optional next-best-view $v^\star$}

\textbf{1. TSDF Initialization:} Fuse depth frames into a coarse TSDF and extract initial surface $\hat{\mathcal{S}}_0$. ;

\textbf{2. Narrow-Band Activation:} Build a sparse voxel hierarchy around $\hat{\mathcal{S}}_0$. ;

\textbf{3. Observation Construction:} Convert depth pixels into samples ${(y_i,\mathbf{a}_i,\sigma_i^2)}$. ;

\textbf{4. Posterior Assembly:}
Compute precision matrix
$\mathbf{Q} \leftarrow \mathbf{Q}_0 + \mathbf{A}^\top \mathbf{W}\mathbf{A}$. ;

\textbf{5. MAP Inference:}
Solve $\mathbf{Q}\boldsymbol{\mu} = \mathbf{h}$ using PCG. ;

\textbf{6. Uncertainty Estimation:}
Approximate marginal variances
$\hat{\mathbf{s}} \approx \mathrm{diag}(\mathbf{Q}^{-1})$ using $K$ probes. ;

\textbf{7. Surface Extraction:}
Extract mesh $\hat{\mathcal{S}}$ via Marching Cubes or Dual Contouring. ;

\textbf{8. Next-Best-View (Optional):}
Select viewpoint
$v^\star = \arg\max_v \mathcal{U}(v)$. ;

\caption{BayesFusion SDF pipeline}
\label{alg:bayesfusion}
\end{algorithm}

\section{Experimental Setup}
\subsection{Data and Baselines}
Results are reported on (i) a controlled scene used for systematic ablations (with a reference surface for evaluation), and (ii) one CO3D object-sequence case study~\cite{reizenstein2021co3d}. Comparisons are made against the TSDF bootstrap output produced by the same pipeline, and a TSDF mesh baseline for the CO3D sequence.

\subsection{Geometry Metrics}
Chamfer distance (L2), mean accuracy, mean completeness, and F-score at multiple distance thresholds are reported.
Suppose $\mathcal{P}$ be the predicted surface points and $\mathcal{G}$ be the ground truth points:
\begin{equation}
\mathrm{CD}(\mathcal{P},\mathcal{G})
=
\frac{1}{|\mathcal{P}|}\sum_{\bm{p}\in\mathcal{P}} \min_{\bm{g}\in\mathcal{G}} \|\bm{p}-\bm{g}\|_2^2
+
\frac{1}{|\mathcal{G}|}\sum_{\bm{g}\in\mathcal{G}} \min_{\bm{p}\in\mathcal{P}} \|\bm{g}-\bm{p}\|_2^2.
\end{equation}

\section{Results}

\subsection{Controlled Ablation Scene}

A controlled environment designed for isolating the contribution of the probabilistic fusion and the TSDF anchoring has been presented in the table~\ref{tab:mainresults}. The anchored BayesFusion-SDF achieved the lowest Chamfer Distance (CD) with the highest F-score at 20~mm, demonstrating the constant improvement in the TSDF bootstrap baseline in the geometric accuracy and the surface coverage. Thus, the removal of the anchor improves the completeness despite significant drop in the accuracy indicating that the TSDF anchoring provides an crucial geometric prior to stabilized inference near surface as described in table~~\ref{tab:mainresults}.

\begin{table}[h]
\centering
\caption{Controlled-scene quantitative results}
\label{tab:mainresults}
\begin{tabular}{lccccc}
\toprule
Method & CD$\downarrow$ & Acc$\downarrow$ & Comp$\downarrow$ & F@20$\uparrow$ & F@50$\uparrow$ \\
\midrule
TSDF bootstrap & 0.00458 & 0.02070 & 0.03617 & 0.3790 & 0.9429 \\
no anchor BayesFusion--SDF & 0.00407 & 0.04115 & 0.01880 & 0.3992 & 0.7670 \\
BayesFusion--SDF+anchor & \textbf{0.00373} & 0.02206 & 0.02412 & \textbf{0.6532} & 0.9365 \\
\bottomrule
\end{tabular}
\end{table}

\begin{figure}[ht]
\centering
\begin{minipage}{0.5\linewidth}
\centering
\includegraphics[width=\linewidth]{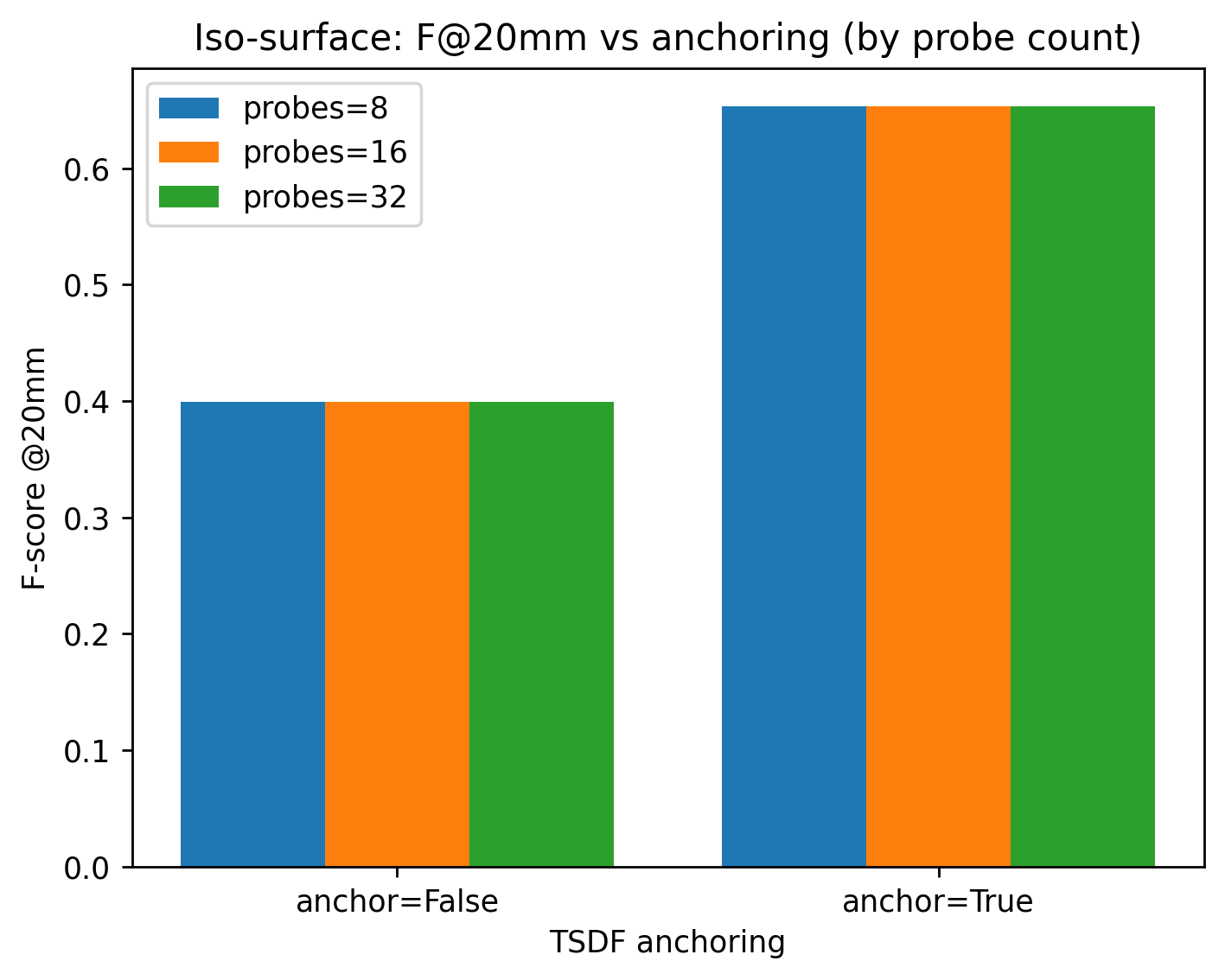}\
\vspace{-0.1em}\small (a) F@20mm vs anchor
\end{minipage}\hfill
\begin{minipage}{0.5\linewidth}
\centering
\includegraphics[width=\linewidth]{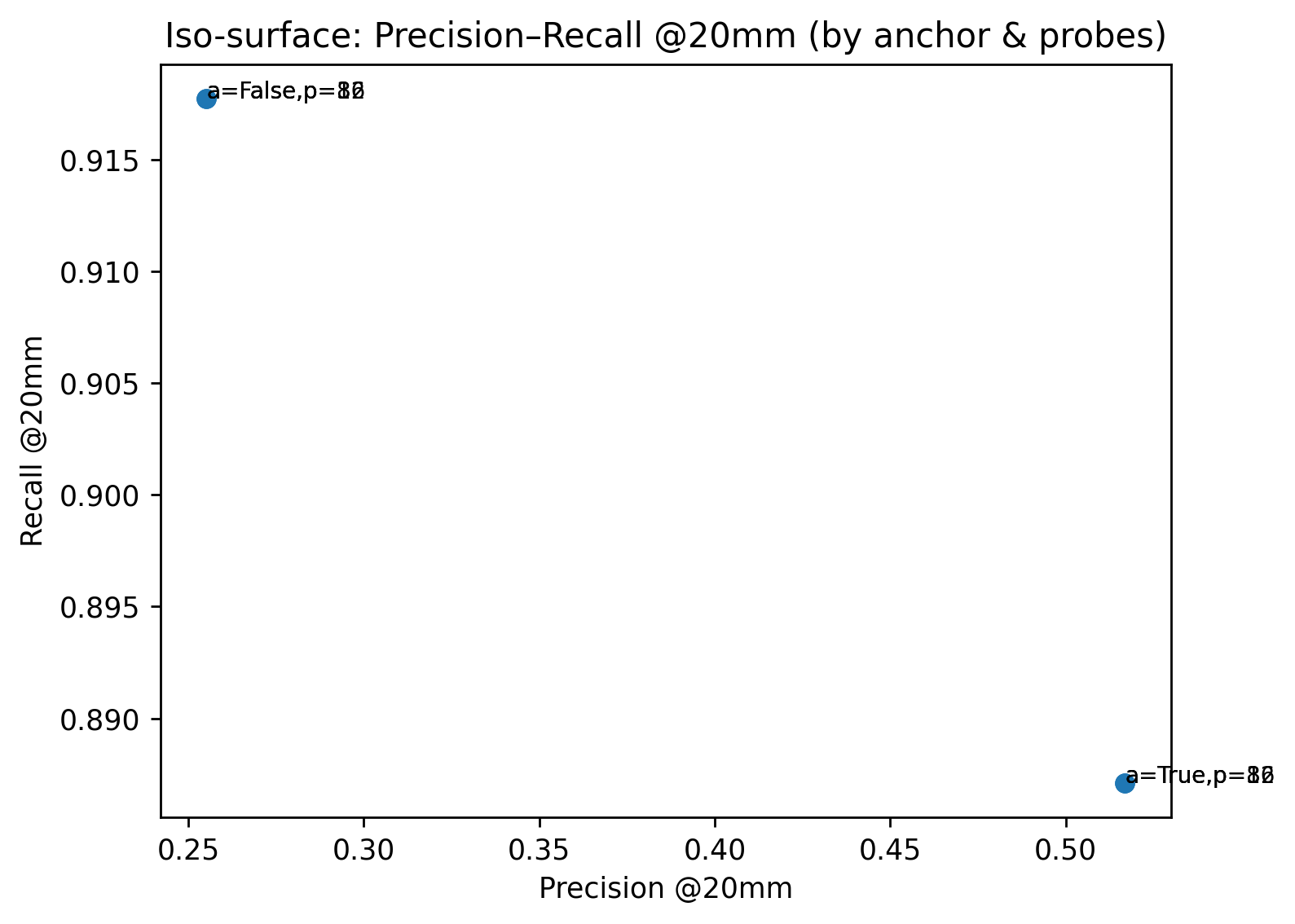}\
\vspace{-0.1em}\small (b) 20mm Precision vs recall
\end{minipage}
\vspace{-0.3em}
\caption{TSDF Anchoring Effect}
\label{fig:anchor_effect}
\end{figure}

The visual effect of anchoring on the precision recall behavior and F-score patterns can be observed in figure~\ref{fig:anchor_effect} which shows the improvement at modern thresholds while maintaining the competitive precision and describes large improvement in F@20.

\subsection{NBV Utility Trends}

\begin{figure}[h]
\centering
\includegraphics[width=0.75\linewidth]{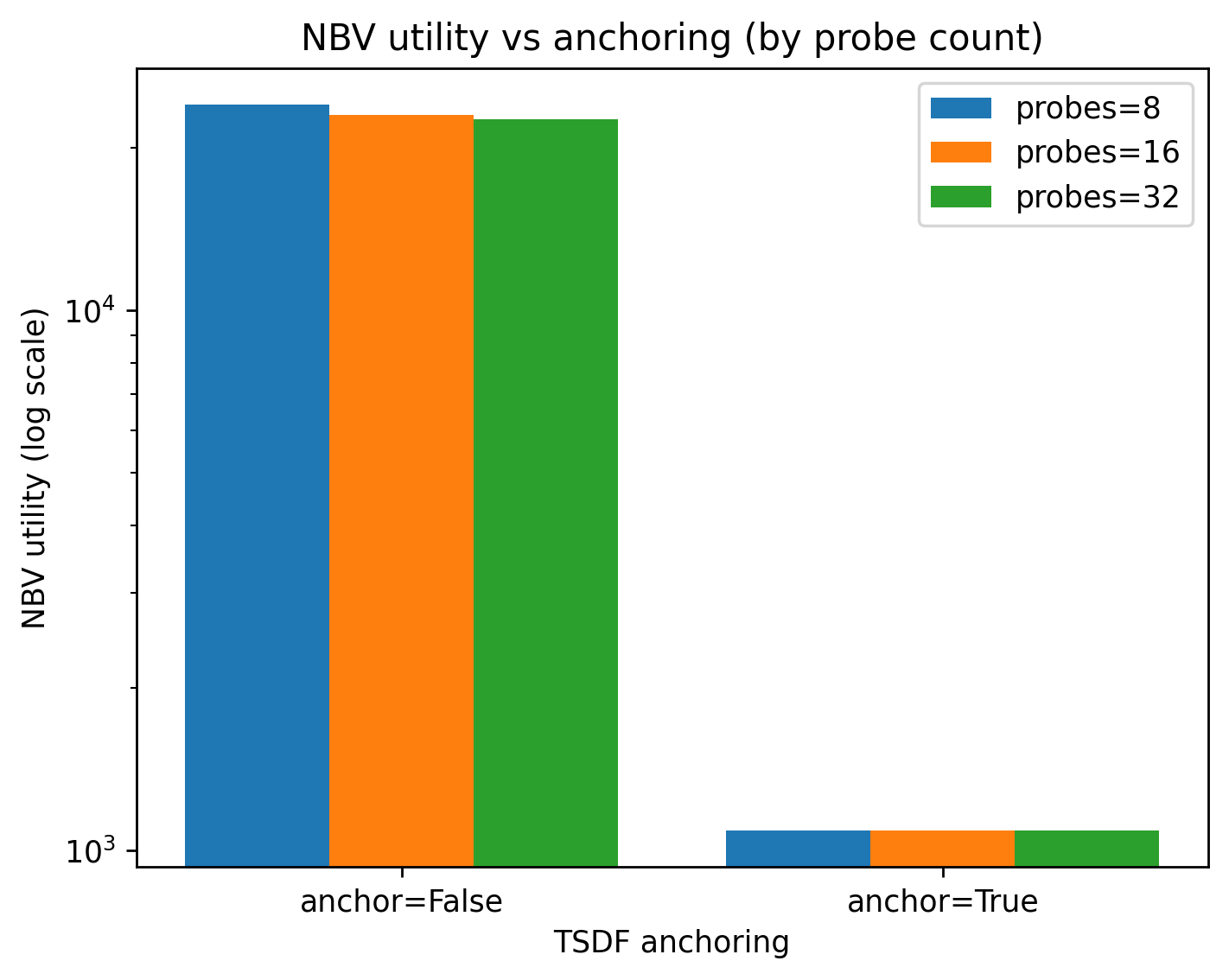}
\vspace{-0.5em}
\caption{NBV Utility Trends}
\label{fig:nbv_utility}
\end{figure}

The utility of the next-best-view (NBV) measured is the expected variance reduction under anchoring configurations as shown in figure~\ref{fig:nbv_utility}, with the anchored formulation producing consistent higher utility probe settings, indicating that the TSDF priors concentrate the informative regions and improve view selection effectiveness.

\subsection{CO3D Case Study}

A performance evaluation has been conducted on real--world data from the CO3D dataset as described in the table~\ref{tab:co3d}. The proposed shows improvement in Chamfer Distance (CD) and completeness during evaluation on iso--surface compared to that of the TSDF baseline, indicating better geometry recovery in difficult conditions.

\begin{table}[h]
\centering
\caption{CO3D sequence}
\label{tab:co3d}
\begin{tabular}{lcccc}
\toprule
Method & CD$\downarrow$ & Acc$\downarrow$ & Comp$\downarrow$ & F@20$\uparrow$ \\
\midrule
TSDF mesh (base) & 0.1195 & 0.1110 & 0.2436 & 0.01509 \\
Ours iso points & \textbf{0.1089} & 0.1139 & \textbf{0.2278} & 0.00994 \\
Ours iso mesh & 0.1231 & \textbf{0.1094} & 0.2500 & 0.01128 \\
\bottomrule
\end{tabular}
\end{table}

The experiment on the proposed BayesFusion–-SDF describes a positive improvement on the TSDF baseline in controlled conditions, showing positive performance in a real--world dataset while providing principled NBV planning.

\section{Discussion and Limitations}

The proposed framework shows that uncertainty-aware reconstruction for CPU-oriented volumetric pipelines is possible by adding probabilistic inference to TSDF fusion. The GMRF method interpreted posterior estimation by directly integrating facilities with NBV objectives. The experimental observation demonstrates the TSDF anchoring stability and geometric precision in the area of sparse measurements while estimating actionable insights for view selections.

Despite the advancements, some limitations can be observed, as the probabilistic formulation can add more memory consumption, which increases the standard TSDF fusion because of building sparse linear systems and iterative solutions. Despite narrow band processing and sparse data structures lowering the cost, solver time, and memory bandwidth, limitation scalability to higher resolutions and large environments. Additionally, using random probing estimates the approximate posterior variance, which is needed to solve multiple linear problems, and makes the program run longer at larger areas of uncertainty, which adds discretization threshold sensitivity in isosurfaces, affecting the F-score even if CD or completeness improved. 

The requirement of enhancing the solver methods for scalability, selection of adaptive parameters, prior learning for geometry regularization, and additional seamless integration of autonomous exploration systems needs to be experimented with in the future for dynamic real-world reconstruction environments.

\section{Conclusion}

The paper introduces a probabilistic signed distance fusion framework designed specially for CPU-only dense 3D reconstruction named BayesFusion--SDF, which features the explicit uncertainty estimation and active view planning capabilities, adding the formulation of SGMRF to the classical TSDF initialization. The method utilizes the principal Bayesian inference of the signed distance values while keeping the computational scalability through the linear algebra and the iterative solvers. The estimation of random diagonals makes it easier to estimate posterior variance close to the reconstruction of the surface that is useful and actionable output instead of adding extra numbers.

The experimental results show the additional probabilistic inference and TSDF anchoring to the geometric reconstruction making it better than the used NBV reasoning. The CPU-only inferencing design keeps it simple, deployable, and compatible with robotics in resource-limited environments with problems that can be GPU-based neural reconstruction pipelines. 

Despite a few limitations that can be observed, such as probabilistic formulation adding more memory and processing power than traditional TSDF fusion, especially when trying to figure out how uncertain things are above large areas, and the performance depending on how the parameters are chosen. The future enhancement of the methods will be the expansion of the framework to encompass dynamic scenes and the incorporation of the learning prior to the enhancement of the geometric fidelity to investigate more robust interconnection between the probabilistic reconstruction and the active perception policies.

The proposed framework made it possible to make uncertainty-aware dense reconstruction with active view planning in the deterministic CPU-based pipeline in a different approach in comparison to the classical volumetric fusion and neural implicit reconstruction.

\bibliographystyle{plain}  

\section*{Author contributions}

\textbf{Soumya Mazumdar} came up with the methodology, wrote the core algorithms, ran experiments, came up with theories, tested them, and wrote the first draft of the manuscript. \textbf{Vineet Kumar Rakesh} helped with the design of the system and the editing of the manuscript. \textbf{Tapas Samanta} independently verified the experimental results and analyses to ensure technical accuracy and consistency. All of the authors looked over and approved the final draft.

\section*{Acknowledgments}

The Variable Energy Cyclotron Centre (VECC), the Department of Atomic Energy (DAE), and the Government of India (GoI) all helped with this work by giving us the tools and technical support we needed to do this research. The authors would like to thank the peer reviewers for their helpful comments and suggestions, as well as the staff of the VECC library for their help during this study.

\section*{Declaration of Competing Interest}

All authors assert that they have no financial or personal connections that could be construed as affecting the work presented in this study. There are no conflicts of interest.

\section*{Data Availability Statement}

The datasets utilized in this study comprise publicly accessible resources and controlled experimental data produced during the research. You can get the CO3D dataset that was used for testing from its official repository. The corresponding author will provide additional experimental results and implementation details upon reasonable request.

\section*{Author Biographies}

\AuthorBio{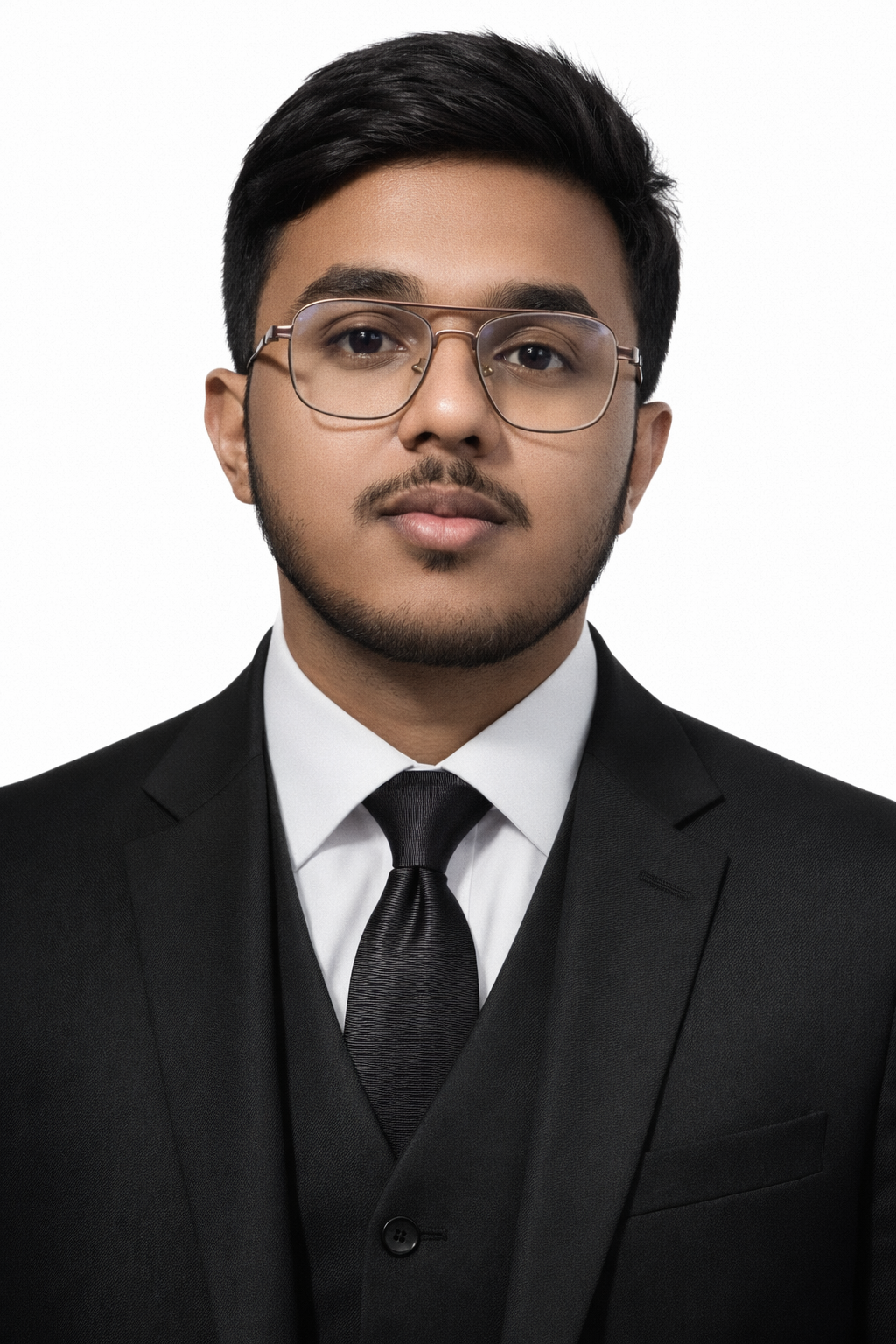}{Soumya Mazumdar}{
is pursuing a dual degree: a B.Tech in Computer Science and Business Systems from Gargi Memorial Institute of Technology, and a B.S. in Data Science from the Indian Institute of Technology Madras. He has contributed to interdisciplinary research with over 25 publications in journals and edited volumes by Elsevier, Springer, IEEE, Wiley, and CRC Press. His research interests include artificial intelligence, machine learning, 6G communications, healthcare technologies, and industrial automation.
}

\AuthorBio{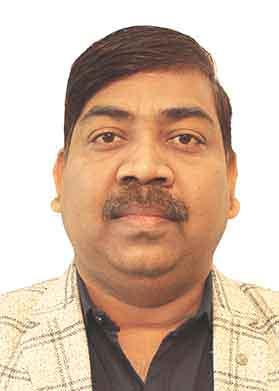}{Vineet Kumar Rakesh}{
is a Technical Officer (Scientific Category) at the Variable Energy Cyclotron Centre (VECC), Department of Atomic Energy, India, with over 22 years of experience in software engineering, database systems, and artificial intelligence. His research focuses on talking head generation, lipreading, and ultra-low-bitrate video compression for real-time teleconferencing. He is pursuing a Ph.D. at Homi Bhabha National Institute, Mumbai. Mr. Rakesh has contributed to office automation, OCR systems, and digital transformation projects at VECC. He is an Associate Member of the Institution of Engineers (India) and a recipient of the DAE Group Achievement Award.
}

\AuthorBio{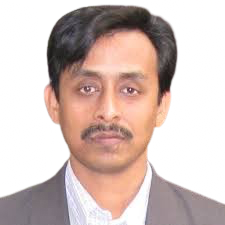}{Dr. Tapas Samanta}{
is a senior scientist and Head of the Computer and Informatics Group at the Variable Energy Cyclotron Centre (VECC), Department of Atomic Energy, India. With over two decades of experience, his work spans artificial intelligence, industrial automation, embedded systems, high-performance computing, and accelerator control systems. He also leads technology transfer initiatives and public scientific outreach at VECC.
}

\end{document}